\documentclass[journal]{IEEEtran}
\ifCLASSINFOpdf
\else
\fi
\usepackage{amsmath}
\usepackage{epsfig}
\usepackage{graphicx}
\usepackage{xcolor}
\usepackage{amssymb}

\usepackage{hyperref}
\hypersetup{
  linkcolor  = red,
  citecolor  = green,
  urlcolor   = blue,
  colorlinks = true,
}

\begin{document}
%
\title{DeepSWIR: A Deep Learning Based Approach for the Synthesis of Short-Wave InfraRed Band using Multi-Sensor Concurrent Datasets}
%
%
%

\author{Litu~Rout,
        Yatharath~Bhateja, Ankur~Garg, Indranil~Mishra, S Manthira~Moorthi, and Debjyoti~Dhar
\thanks{All authors are with the Optical Data Processing Division, Signal and Image Processing Group, Space Applications Center, Indian Space Research Organisation, Ahmedabad, India - 380015.}
\thanks{Corresponding mail ids: (lr, ybhateja, agarg, indranil, smmoorthi, deb)@sac.isro.gov.in}}%
\maketitle

\begin{abstract}
Convolutional Neural Network (CNN) is achieving remarkable progress in various computer vision tasks. In the past few years, the remote sensing community has observed Deep Neural Network (DNN) finally taking off in several challenging fields. In this study, we propose a DNN to generate a predefined High Resolution (HR) synthetic spectral band using an ensemble of concurrent Low Resolution (LR) bands and existing HR bands. Of particular interest, the proposed network, namely DeepSWIR, synthesizes Short-Wave InfraRed (SWIR) band at 5m Ground Sampling Distance (GSD) using Green (G), Red (R) and Near InfraRed (NIR) bands at both 24m and 5m GSD, and SWIR band at 24m GSD. To our knowledge, the highest spatial resolution of commercially deliverable SWIR band is at 7.5m GSD. Also, we propose a Gaussian feathering based image stitching approach in light of processing large satellite imagery. To experimentally validate the synthesized HR SWIR band, we critically analyse the qualitative and quantitative results produced by DeepSWIR using state-of-the-art evaluation metrics. Further, we convert the synthesized DN values to Top Of Atmosphere (TOA) reflectance and compare with the corresponding band of Sentinel-2B. Finally, we show one real world application of the synthesized band by using it to map wetland resources over our region of interest.

\end{abstract}

\begin{IEEEkeywords}
Remote Sensing, Convolutional Neural Network, Band Synthesis, Super Resolution, Image Stitching.
\end{IEEEkeywords}

%
\IEEEpeerreviewmaketitle

\section{Introduction}
%
%
%
%

\begin{figure*}[!t]
	\centering
	\includegraphics[scale=0.6]{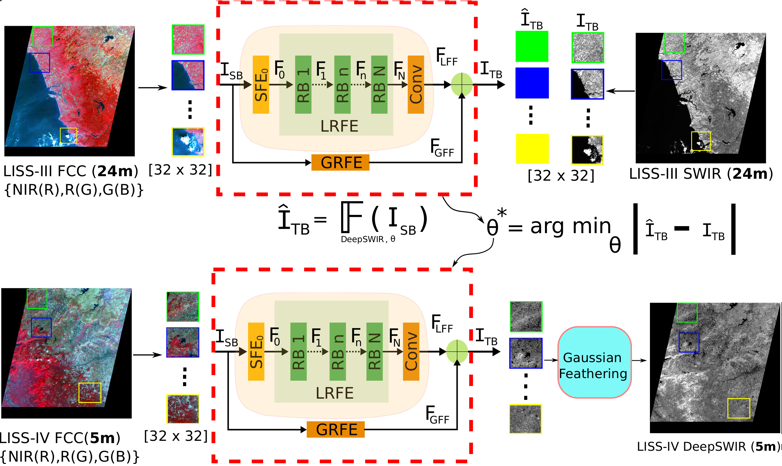}
	\caption{Pipeline of the proposed framework. The optimal set of trainable parameters, $\theta^{*}$ is obtained by training the DeepSWIR model end to end on the low resolution concurrent multi-spectral bands of LISS-III. The learned set of parameters is  then used to synthesize a virtual high resolution band, $SWIR_{5m}$ by leveraging the spectral characteristics from LISS-III and spatial characteristics from LISS-IV.} 
	\label{framework}
\end{figure*}

\IEEEPARstart{N}{umerous} Remote Sensing (RS) applications have a strong reliance on earth observation images which are highly resolved in spatial domain and preserve the spectral characteristics of individual objects~\cite{patino2013review,pohl1998review,blaschke2010object,colomina2014unmanned}. In the recent decade, several optical and Synthetic Aperture Radar (SAR) satellites were launched with high spatio-spectral resolution that opens unprecedented opportunities in applications including Object Detection~\cite{cheng2016survey}, Image Retrieval~\cite{yang2013geographic}, Automatic Target Recognition~\cite{wagner2016sar}, Semantic Segmentation~\cite{volpi2015semantic}, Terrain surface classification~\cite{geng2017deep}, detection of archaeological features in modern landscapes~\cite{tapete2017trends}, biomass assessment~\cite{kumar2016above}, identification of forest cutting~\cite{khati2018identification}  etc. The extrapolation of Visible  and Near InfraRed (VNIR) bands farther into InfraRed (IR) spectrum offers the provision to capture rich features uniquely associated with applications such as, material identification, wildfire response, mining/geology etc. Due to minimal influence of atmospheric noise, fog, smoke etc. on SWIR band, this band is relatively more suitable for the aforementioned applications compared to VNIR bands.

 Despite the usefulness of SWIR band, very few RS satellites are capable of providing high resolution spectral bands in this particular range of the Electro Magnetic Spectrum (EMS). To our knowledge, the highest spatial resolution of commercially deliverable SWIR band is at 7.5m, which is provided by WorldView-3~\cite{ye2017assessment}. Of particular interest, the Indian Remote Sensing (IRS) satellite, Resourcesat-2A provides VNIR and SWIR at 24m spatial resolution using Linear Imaging and self Scanning Sensor (LISS-III). However, the LISS-IV sensor present in Resourcesat-2A provides only VNIR bands at 5m spatial resolution. For this reason, generation of a synthetic High Resolution (HR) SWIR (5m) band by preserving the spectral characteristics of Low Resolution (LR) SWIR (24m), beyond naive interpolation~\cite{anbarjafari2010image} and data fusion~\cite{thomas2008synthesis}, is the primary motive of this study. To tackle this problem, we propose a customized deep neural network (DeepSWIR) by borrowing the integral parts from the state-of-the-art CNN architectures~\cite{krizhevsky2012imagenet,szegedy2015going,he2016deep,huang2017densely}, which is trained on only the LR concurrent bands of LISS-III (24m) and specifically designed to synthesize LISS-IV-SWIR at 5m spatial resolution.
 
\begin{figure*}[!th]
	\centering
	\includegraphics[scale=0.6]{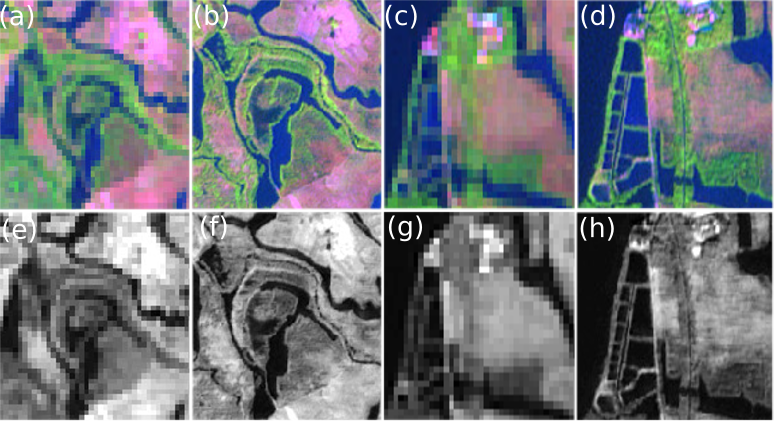}
	\caption{Synthesis of $SWIR_{5m}$ by leveraging spatial and spectral characteristics from $\left \{ G,R,NIR \right \}_{5m}$ and $\left \{ G,R,NIR,SWIR \right \}_{24m}$, respectively. (a,b) and (c,d) represent the paired False Colour Composite (FCC) of LISS-III (24m) $\left (  SWIR~(R), NIR~(G), R~(B) \right )$ and LISS-IV (5m) $\left (  DeepSWIR~(R), NIR~(G), R~(B) \right )$ satellite imagery. (e,f) and (g,h) represent their corresponding $SWIR$ bands. The FCC verifies that the synthesized $SWIR_{5m}$ is auto-registered with existing LISS-IV bands and hence, maintains the spatial resolution. The paired $SWIR_{24m}$ and $SWIR_{5m}$ assures that the spectral characteristics is preserved.} 
	\label{main_l3l4}
\end{figure*}

 A vast majority of the RS community is addressing the issue of generating HR spectral bands by posing it as a super resolution problem and hallucinating the missing HR details in the LR bands. In the context of the current scenario, a similar objective would be to super-resolve the LISS-III bands at 24m GSD to the corresponding LISS-IV bands at 5m GSD, thereby meeting our objective of synthesizing the required spectral band, i.e., $SWIR_{5m} $. Thus, in order to construct training data for such supervised learning, both LR and HR bands corresponding to the same region on ground and at the same time is necessary, which is not practically viable even with sophisticated technology. Especially, this approach, without any assumption, is not feasible at present for faithful super-resolution of our band of interest due to unavailability of $SWIR_{5m} $. However, a practically viable way of solving the problem, as proposed by Charis et al.~\cite{lanaras2018super}, is to assume that the mapping from LR to HR bands is roughly scale-invariant and it depends only on the relative difference in resolution, i.e., $\left\lbrace 20m \right\rbrace \rightarrow \left\lbrace 10m \right\rbrace \equiv \left\lbrace 40m \right\rbrace \rightarrow \left\lbrace 20m \right\rbrace$. Thus, virtually unlimited data can be prepared by downsampling our 24m bands to a required level and learning how to transfer necessary HR details from the concurrent LR bands at required level to the existing bands at 24m. Then the trained model is applied to bands at 24m and expected to improve the resolution by the same factor as from the required level to 24m. Nevertheless, the process of constructing training data in such manner involves downsampling, filtering (e.g. Gaussian blurring) etc. that inherently injects noise into the training data. Though this method is quite successful in learning the inverse mapping,  this reduces the scientific correctness necessary for our application with required degree of realism. Therefore, we propose to learn the spectral mapping directly from real bands of LISS-III $\left\lbrace G,R,NIR \right\rbrace_{24m} \rightarrow \left\lbrace SWIR \right\rbrace_{24m}$, while preserving the spatial characteristics. Then we feed the existing HR bands of LISS-IV $\left\lbrace G,R,NIR \right\rbrace_{5m}$ to the trained model and expect a virtual band that establishes the learned LR spectral mapping while preserving the input spatial resolution, i.e., 5m GSD. The underlying hypothesis is that the model tries to establish a robust spectral mapping of various objects as a function of concurrent LR bands in the hidden layers of a very deep neural network which does not undergo drastic changes across spatial resolution. Fig.~\ref{framework} illustrates the pipeline of our overall framework.  Fig.~\ref{main_l3l4} pictorially depicts the qualitative assessment of proposed HR band synthesis using our customized DeepSWIR.

Our contributions are loosely summarized as three-folds. 
\begin{itemize}
\item First, we propose a unified deep learning based framework for synthesizing HR required spectral band(s) using LR and existing HR concurrent bands. The proposed model, namely DeepSWIR, leverages the spectral characteristics from LR bands and spatial, from HR bands.

\item In this process, we propose an architecture to perform both global and local residual learning in order to fill in the missing high frequency details of LR band. Thus, the combined residual learning allows adaptive fusion of deep and shallow features for the synthesis of required band.

\item Further, we propose a simple and yet, effective approach for processing large satellite images. By this approach, we process large number of overlapping crops ([32x32]) of a large satellite image and stitch them together using our Gaussian feather mosaicing scheme for seamless band generation.
\end{itemize}

In the following Section~\ref{Related}, we briefly discuss various approaches intended to address the problem similar to ours, i.e., generating required HR spectral band(s) using LR bands and existing HR bands. Thereafter, we detail the proposed  methodology in Section~\ref{Proposed} and the experimental details in Section~\ref{Experiments} to critically analyse the performance of the proposed DeepSWIR. At the end , we draw compelling inferences based on our discussion throughout and discuss concluding remarks in Section~\ref{Conclusion}.

\section{Related Work}
\label{Related}
The super-resolution enhancement techniques have gradually developed from naive interpolation~\cite{zhang2006edge} to CNN based supervised~\cite{dong2014learning} and unsupervised approach~\cite{ledig2017photo}. A popular class of conventional algorithms, which are designed to address similar kind of RS applications, attempt to enhance the spatial resolution of the LR bands by fusing the high frequency details from Panchromatic (Pan) images. The smaller size of pixels that leads to high spatial resolution in Pan images is the primary contributing factor in image fusion. This method of pan-sharpening, however, is affected by the possibilities of dis-similar modalities caused by different spectral and temporal acquisition even after geometric registration. In addition, the lack of Pan bands in Resourcesat-2A and unavoidable shortcomings of fusion based schemes leads to recommendations for developing a new technique for our application. \\

As per recent studies~\cite{zhang2017image,zhang2018density,zhang2018densely}, the CNN based approaches offer considerable gain over bicubic and sophisticated neighbourhood regression methods on non-remote sensing benchmarks such as CelebA~\cite{escalera2016chalearn} or ImageNet~\cite{deng2009imagenet} in similar tasks. Dong et al.~\cite{dong2014learning} proposed Super-Resolution Convolutional Neural Network (SRCNN) that outperformed the state-of-the-art methods till date by a large margin.  Among generative models, Chrisitian et al.~\cite{ledig2017photo} proposed the Photo-Realistic Single Image Super-Resolution using a Generative Adversarial Network (SRGAN), which pushed the synthetic images towards natural image manifold that made it perceptually more convincing. Kim et al.~\cite{kim2016deeply} discussed the idea of recursive or shared weights in Deeply-Recursive Convolutional Network (DRCN), which performed favourably against state-of-the-art methods with relatively fewer parameters. Mao et al.~\cite{mao2016image} proposed a very deep encoder-decoder architecture with residual connections, which performed better than SRCNN in super resolution tasks. Shi et al.~\cite{shi2016real} developed the Efficient Sub-Pixel Convolutional Neural Network (ESPCN) that increased resolution in the last layer of ESPCN, thereby reducing the computational and space complexity by a large margin. Jin et al.~\cite{yamanaka2017fast} proposed a Deep CNN with Skip Connection and Network in Network (DCSCN), which extracts cues from both local and global area by leveraging CNN with skip connections and Network in Network architecture (NIN)~\cite{lin2013network}.\\

Among RS satellite imagery, Charis et al.~\cite{lanaras2018super} integrated residual blocks in a customized CNN, namely DSen2, to generate super-resolved imagery of Sentinel2 at 10m GSD. This method, however, operates under the assumption of scale-invariance which diverges from our objective of learning spectral mapping from the existing bands directly, while preserving the spatio-spectral characteristics. To the contrary, the proposed deep learning model (DeepSWIR), unlike DSen2, remains less susceptible to the adverse effects of inherent noise injection into the training data as a result of resampling, mainly due to the use of existing bands in training DeepSWIR. Thus, deep learning being a data driven process, relatively more appropriate information regarding the spatial and spectral characteristics of individual objects flow from the LR LISS-III data to the DeepSWIR model, which is transferred to the HR LISS-IV data without loss of generality. Darren et al.~\cite{pouliot2018landsat} assimilated the effectiveness of shallow and deep CNN for super-resolution tasks on satellite imagery. As per their exploratory analysis, the SRCNN~\cite{dong2014learning} with deep connectivity and residual connections (DCR\_SRCNN)~\cite{pouliot2018landsat}  provided the best results across their regions of interest. In this study, we observed a similar trend in performance with respect to depth of the network, and hence, we attempted to implement a deeper network borrowing the concepts from few of the latest developments in this area~\cite{krizhevsky2012imagenet,szegedy2015going,he2016deep,huang2017densely}.

\section{Proposed Methodology}
\label{Proposed}
Here, we detail our method by expanding the fundamental components of the end-to-end framework. At first, we briefly discuss about the satellite imagery under the scope of our study in Section \ref{Preparation}. Thereafter, we describe about datasets used in training and cross-validation in Section~\ref{data}, the customized DeepSWIR architecture in Section~\ref{Architecture}, and image stitching mechanism using Gaussian weights in Section~\ref{feather}.

\subsection{Resourcesat-2A Satellite Imagery}
\label{Preparation}
The proposed DeepSWIR exploits the similarity and concurrency associated with the unique sensors: LISS-III and LISS-IV on board Indian Space Research Organisation's (ISRO) Resourcesat-2A mission. In particular, we take advantage of same wavelength range in the VNIR part of the EMS and thus, create a representative dataset for training the model without altering the pure spectra of the acquired bands. Due to simultaneous acquisition, the LISS-IV sensor's 70km swath overlaps with the 140 km swath of the LISS-III sensor, thereby providing an unique opportunity to create concurrent datasets under near identical environmental condition~\cite{pandya2013quantification}.

\begin{table}[!th]
\centering
\caption{LISS-III and LISS-IV sensor characteristics}
\label{band_specs}
\begin{tabular}{|l|l|l|}
\hline
Bands and Wavelength Ranges $ (\mu m)$        & LISS-III  & LISS-IV \\ \hline \hline
Band2 (G)    & 0.52-0.59 & 0.52-0.59 \\ \hline
Band3 (R)    & 0.62-0.68 & 0.62-0.68 \\ \hline
Band4 (NIR)  & 0.77-0.86 & 0.77 -0.86 \\ \hline
Band5 (SWIR) & 1.55-1.70 &   -        \\ \hline \hline
Swath Width (km) & 140 &   70        \\ \hline
Spatial Resolution (m) & 24 &  5        \\ \hline
Radiometric Resolution (bit) & 10 &  10        \\ \hline
Temporal Resolution (days) & 24 &  48     \\ \hline
\end{tabular}
\end{table}

The Resourcesat-2A satellite orbits in a sun-synchronous orbit at an altitude of 817 km. The satellite completes about 14 orbits per day with 101.35 minutes to complete one revolution around the earth. To cover the entire earth, it takes around 341 orbits during a 24-day cycle. Both LISS-III and LISS-IV have VNIR bands in the same range of EMS, but SWIR band is present only in LISS-III sensor. The LISS-III sensor covers a 140 km swath with deliverable product at 24m spatial resolution, whereas the LISS-IV sensor covers a 70 km swath with 5m spatial resolution. Both LISS-III and LISS-IV have same radiometric resolution of 10 bit radiometry. The characteristics of LISS-III and LISS-IV sensors are given in Table \ref{band_specs}~\cite{pandya2013quantification}.

\subsection{Preparation of Representative Dataset}
\label{data}

\begin{figure*}[!th] 
	\begin{center}
	\includegraphics[width=0.9\textwidth]{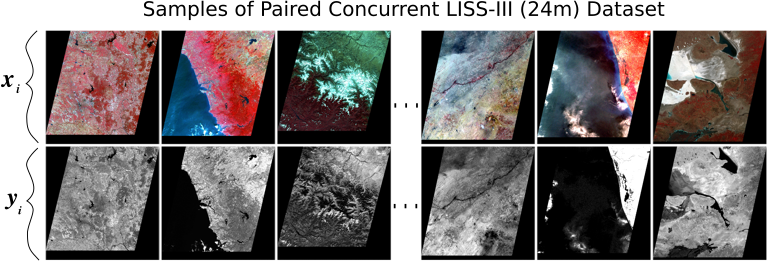}
	\caption{Samples of dataset used in training and cross validation of DeepSWIR. Here, $x_i$ represents the False Colour Composite (FCC) of $\left \{ NIR (R), Red(G), Green (B) \right \}_{24m}$ and $y_i$ represents the corresponding concurrent $SWIR_{24m}$. }
	\label{sample}
	\end{center}
\end{figure*}

One of the major challenges in supervised deep learning algorithms is the preparation of paired representative datasets to cover the area of interest. As our area of interest is limited to different regions of India, we choose few representative path \& row combinations as per Worldwide Reference System (WRS), which essentially covers different unique object signatures in SWIR range of the EMS. Of particular interest, the training dataset is constructed to learn the spectral characteristics of urban area, water bodies (lake, river, ocean etc.), vegetation, barren land, snow, ice, cloud, desert sand, white sand etc. For validation and testing purpose, a similar representative dataset is contructed to analyze the network's generalization performance across seasonal variation, different path \& row combinations and different spatial resolutions. The Date of Pass (DOP), path, row and area covered of the tiles used in training, validation and testing of DeepSWIR are presented in Table~\ref{data_used} and a sample subset data is shown in Fig.~\ref{sample}.

\begin{table*}[!th]
\caption{Statistics of the data used for training, validation and testing the proposed DeepSWIR.}
\label{data_used}
\centering
\begin{tabular}{|l|l|l|l|l|}
\hline
       &          & Train &     &              \\ \hline
S. No. & DOP      & Path  & Row & Major Features (Area) \\ \hline
1      & 02.03.18 & 93    & 56  & Urban, Land, Vegetation (Ahmedabad)    \\ \hline
2      & 19.01.18 & 99    & 67  & Water, Vegetation,  Land (Kerala)       \\ \hline
3      & 08.01.18 & 91    & 55  & Desert Sand, White Sand (Rajasthan)    \\ \hline
4      & 05.10.17 & 97    & 49  & Snow, Ice, Cloud, Mountain (Himalaya)     \\ \hline
5      & 29.01.17 &  100 & 58  & Vegetation, Water (Madhya Pradesh)     \\ \hline
6      &  11.01.18 & 107   & 52 &  Mountain, Snow, Vegetation(Sikkim)    \\ \hline \hline
       &          & Test  &     &              \\ \hline
S. No. & DOP      & Path  & Row & Major Features (Area) \\ \hline       
1      & 17.11.17 & 96    & 62  & Wetland, Water, Vegetation(Goa)          \\ \hline
2      & 26.11.17 & 93    & 56  & Urban, Seasonal Variation (Ahmedabad)    \\ \hline
3      & 24.11.17 & 107   & 52  & Mountain, Cloud, Snow, Vegetation (Sikkim)       \\ \hline
4      &  14.04.17        &  91     &  52   & Desert  Sand(Jaisalmer)    \\ \hline
\end{tabular}%
\end{table*}

\subsection{Network Architecture}
\label{Architecture}
\begin{figure}[!th]
	\centering
	\includegraphics[width=0.5\columnwidth]{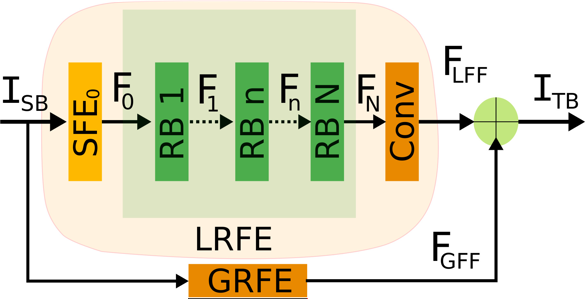}
	\caption{Synthesis of target band ($\text{I}_\text{TB}$) by fusing global ($\text{F}_\text{GFF}$) and local ($\text{F}_\text{LFF}$) features extracted from source bands ($\text{I}_\text{SB}$) with carefully crafted Global Residual Feature Extraction (GRFE) and Local Residual Feature Extraction (LRFE) blocks.}
\label{main}
\end{figure}

The overall architecture of proposed DeepSWIR is shown in Fig.~\ref{main}. As shown in Fig.~\ref{main}, the DeepSWIR model encompasses two major learning mechanisms: global and local residual learning. The local residual learning, as part of a Local Residual Feature Extraction unit ($\mathbb{F}_{LRFE(.)}$), is guided by a Shallow Feature Extraction module $(\mathbb{F}_{SFE}(.))$ and cascaded Residual Blocks $(\mathbb{F}_{RB}(.))$. The global residual learning is undertaken by a Global Residual Feature Extraction unit $(\mathbb{F}_{GRFE}(.))$. Thereafter, a dense fusion of features extracted from local and global residual learning directly predicts the intensities within dynamic range of the required spectral band. 

\begin{figure}[!th]
\begin{center}
	\includegraphics[scale=0.3]{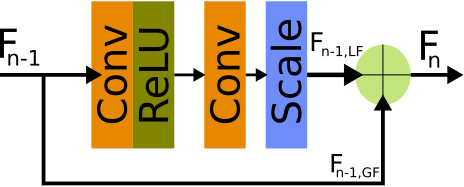}
	\caption{Residual Blocks (RB) used in LRFE.}
	\label{res}    
\end{center}

\end{figure}

\textbf{Local residual learning} operates on the features $\mathbf{F}_{0}$ extracted by the shallow feature extraction layer, 
\begin{equation}
\mathbf{F}_{0} = \mathbb{F}_{SFE}(I_{SB}).
\end{equation}
The $\mathbb{F}_{SFE}(.)$ unit consists of a single convolutional layer without any non-linear activation to extract shallow features. The shallow features are then fed as inputs to the Residual Blocks,
\begin{equation}
\begin{split}
\mathbf{F}_n &= \mathbb{F}_{RB,n}(\mathbf{F}_{n-1}),~n = 1,\dots,N \\
			 &= \mathbb{F}_{RB,n}(\mathbb{F}_{RB,n-1}(\mathbb{F}_{RB,n-2}(\mathbb{F}_{RB,n-3}(\dots \mathbb{F}_{RB,1}(\mathbf{F}_{0}))))).
\end{split}
\end{equation}
Here, $\mathbb{F}_{RB,n}(.)$ denotes the operations of $n^{th}$ residual block inside the LRFE unit. The number of channels present in the internal layers of a residual block is represented by feature size in Section~\ref{ablation}-Table~\ref{ablation_tbl}. As shown in Fig.~\ref{res}, the residual block used in DeepSWIR is a slight variant of the original residual block with skip connections \cite{he2016deep} in a sense that it does not use activation units at the output and is expected to adaptively capture the dynamic range of radiometric values. Hence, the output of $\mathbb{F}_{RB,n}(.)$ is computed by fusing the global features $\mathbf{F}_{GF}$ with the scaled local features $\mathbf{F}_{LF}$,
\begin{equation}
\mathbf{F}_{n} = \mathbf{F}_{n-1,GF} + \mathbf{F}_{n-1,LF}.
\end{equation}

 Thus, the adverse effects which may arise due to deeper convolution is suppressed to some extent through guided residual learning at each stage of the LRFE module. In addition, we verify that the residual block in a deep neural network, as shown in several recent studies~\cite{he2016deep,pouliot2018landsat,yamanaka2017fast}, introduces fast convergence and better information flow. We, therefore, use these blocks to address issues of similar kind.

Further, the final stage of local residual learning consists of a Local Feature Fusion $(\mathbb{F}_{LFF}(.))$ layer to operate on the features extracted from previous layers and fuse them to maintain the channel dimension (here, 1) of the target band $(\mathbf{I}_{TB})$,
\begin{equation}
\mathbf{F}_{LFF} = \mathbb{F}_{LFF}(\mathbf{F}_{N}).
\end{equation}

\textbf{Global Residual Learning}, to the contrary, operates on the source bands $I_{SB}$ and extracts shallower features while preserving higher level structures present in the source imagery. The GRFE unit is built upon fully convolutional layers with kernel size [1x1]. One way to understand the operations of $\mathbb{F}_{GRFE}(.)$ is that it learns to preserve the necessary spatial characteristics of the source imagery. Thereby, it contributes significantly to reconstruct the target band at same spatial resolution as that of the source band without deteriorating the spatial characteristics,  
\begin{equation} 
\mathbf{F}_{GFF} = \mathbb{F}_{GRFE}(\mathbf{I}_{SB}).
\end{equation}

On the other hand, the LRFE unit is expected to contribute mainly towards learning the spectral mapping from source bands to target band. Thus, the required target band is synthesized by combining the spatial and spectral information in the form of global and local residual feature fusion, respectively,
\begin{equation}
\mathbf{I}_{TB} = \mathbf{F}_{GFF} + \mathbf{F}_{LFF}.
\end{equation}

\subsection{Gaussian Feathering for Image Stitching}
\label{feather}
One of the major shortcomings in employing deep learning on satellite imagery is to process large volume of data. Unlike non-remote sensing datasets, it is very unlikely to efficiently process a full satellite image through deep convolutional networks. A common solution to address such problems is to randomly crop several small portions of each individual large tile and train neural networks on this exhaustive cropped dataset. Though this approach reduces the computational complexity to some extent while training, it gives rise to a typical problem while reconstructing a seamless full tile from the predicted crops. In particular, it produces blocky effects while stitching small crops to generate a full tile. To tackle this problem, we crop overlapping patches in row major order and stitch them together in the same order by our feather mosaicing scheme with Gaussian weights. \\

\begin{figure}
    \centering
    \includegraphics[width=0.9\columnwidth]{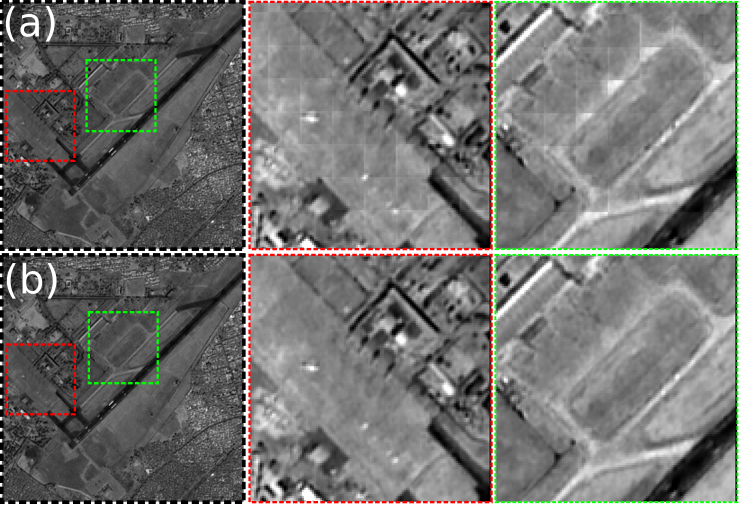}
    \caption{Qualitative analysis of Gaussian feathering for image stitching. (a) Naive stitching and (b) Gaussian feathering of predicted patches. Though blocky effect is not distinctively visible at native resolution, it degrades the data quality as it persists throughout the image. The proposed Gaussian feathering does not suffer from blocky effect even at 4 times upsampling as highlighted in Red and Green bounding boxes.}
    \label{stitch}
\end{figure}

Let $x_i \in \mathbb{R}^{d \times d \times c},~i = 1,2,\dots,N$ represent the ordered crops (row major) of a full tile $X \in \mathbb{R}^{m \times n \times c}$, where $d,~c,~N,~\text{and}~(m,n)$ represent the size of a small patch, number of input channels (bands), total number of crops, and size of input tile, respectively. The crops of target band, $Y\in \mathbb{R}^{m \times n \times 1}$ corresponding to $X$ are synthesized by,
\begin{equation}
y_{i} = \mathbb{F}_{DeepSWIR}(x_{i}),~y_i \in \mathbb{R}^{d \times d \times 1}.
\end{equation}
For 50\%, i.e, d/2 pixel overlap between two adjacent patches in a row, we define Gaussian feathering of $y_i$ and $y_{i+1}$ to construct $y_{i,i+1} \in \mathbb{R}^{d \times (d+\frac{d}{2})}$ by, 

\resizebox{0.96\columnwidth}{!}{
$ y_{i,i+1}(u,v)_{u \in [1,d]} = 
  \begin{cases}
    y_i(u,v) & \text{if $v \in [1,\frac{d}{2})$} \\
    y_i(u,v) * \omega_{i}(v-\frac{d}{2}+1) + y_{i+1}(u,v-\frac{d}{2}+1) * \omega_{i+1}(v-\frac{d}{2}+1) & \text{if $v \in [\frac{d}{2},d)$} \\
    y_{i+1}(u,v-\frac{d}{2}+1) & \text{if $v \in [d,d+\frac{d}{2}]$}. \\
  \end{cases}
$}  
Here, $u \in [1,d],~ \omega_{i} \in \mathbb{R}^{1 \times \frac{d}{2}}$ and $\omega_{i+1} \in \mathbb{R}^{1 \times \frac{d}{2}}$ represent the Gaussian weights associated with $i^{th}$ and $(i+1)^{th}$ patch, respectively, i.e, $\omega_i =  \exp \left ( \frac{-(\hat{d}-\mu)^2}{2* \sigma^2} \right ),~\omega_{i+1} = 1-\omega_{i}~\text{for }~\hat{d} \in [\frac{d}{2}, d]$. The reason for selecting Gaussian weights, instead of linear, or sigmoid weights, is to preserve the source distribution of individual optical image patches which is reasonably assumed to follow Gaussian distribution~\cite{permuter2003gaussian,permuter2006study,storvik2008combination,celik2010image}. In our case, we set $\mu = \frac{d}{2}$ and $\sigma=\frac{d}{4}$ so as to meet the computational demand. Note that we first feather adjacent patches in horizontal direction to construct horizontal strips, $\tilde{y}_{j} \in \mathbb{R}^{d \times n}$, where $j = 1,2,\dots,M$ and $M$ represents total number of horizontal strips. Further, we stitch these horizontal strips ($\tilde{y}$) in vertical direction in the similar manner and thus, reconstruct a seamless full tile without any blocky effect. To maintain continuity at the boundary along horizontal and vertical direction, we simply replace the pixels of equivalent size from the corresponding edges with the final predicted patches. Fig.~\ref{stitch} shows the relative improvement of proposed Gaussian feathering over naively stitching individual patches to form a complete tile. In particular, the proposed Gaussian feathering preserves the source radiometry while eliminating abrupt transition between adjacent patches.\\

\section{Experiments and Analysis}
\label{Experiments}
In this section,  we elaborate on the necessary implementation details in Section~\ref{implementation}, used baseline and state-of-the-art evaluation metrics in Section~\ref{metrics}, ablation study of proposed architecture in Section~\ref{ablation}, evaluation on LISS-III and LISS-IV in Section~\ref{eval3}-\ref{eval4} , cross validation with Sentinel-2B in Section~\ref{sen}, and finally, one real world application of synthesized virtual band in Section~\ref{wetland} .  

\subsection{Implementation Details}
\label{implementation}
For faithful regeneration of the reported experimental results, we provide the exact hyper-parameter settings and implementation details as follows. All the experiments have been conducted on a local machine with 256GB ram, 8GB gpu memory, and i7 processor. We consider patch size of $(32 \times 32)$, i.e. $d=32$. The total number of randomly cropped images from the representative dataset, as discussed in Section~\ref{Preparation}, is approximately  1M. For cross-validation purpose we randomly split the total crops into training ($\approx 800K$) and testing ($\approx 200K$) datasets. We use initial learning rate of $1e-4$ and keep decreasing the rate on plateau with a scheduled decay of $0.004$. To avoid vanishing/exploding gradients, we normalize the gradient norm at each time step to 1 and clip each gradient value to $[-0.5, 0.5]$. As reported by D. P. Kingma and J.L.Ba~\cite{kingma2014adam}, we use $\beta 1 = 0.9$ and $\beta 2 = 0.999$ to update the biased first and second moment estimates, respectively. We use NADAM optimizer~\cite{dozat2016incorporating} for faster convergence and better stability. The $\epsilon$ in NADAM optimizer is set to $1e-8$. The entire pipeline has been developed in python using open-source libraries such as Keras~\cite{chollet2015} with TensorFlow~\cite{tensorflow2015-whitepaper} backend. As reported by Charis et al.~\cite{lanaras2018super} the Mean Absolute Error (MAE) along with Root Mean Square Error (RMSE) performs favourably in satellite imagery. We verify the efficacy of such combination and advocate using MAE as our primary objective function to be minimized and RMSE as our early stopping criteria (Section \ref{metrics}). The upper limit of training epochs is set to 10000. The patience value for early stopping criteria is set as 5 epochs, which is an essential factor to control overfitting to some extent. We observed that deeper networks with smaller kernel size works reasonably better. So we use [3x3] convolutional kernels and 28 layers in this case (Section~\ref{ablation}). We also observed that hour-glass structure like CNN with pooling layers losses high frequency image details due to several downsampling and upsampling involved in the feed-forward process. Therefore, we implemented DeepSWIR in a fully convolutional fashion.

\subsection{Baseline and Evaluation Metrics}
\label{metrics}
As we do not have original $SWIR_{5m}$ to assess the performance of synthesized SWIR band, we intend to analyze the performance by comparing with the closest alternative, i.e. bicubic interpolated $SWIR_{24m}$ at 5m GSD. We primarily choose RMSE as our quantitative evaluation metric and therefore, set this as our early stopping criteria while training the model. The RMSE is computed as following, 
\begin{equation}
RMSE = \sqrt{\frac{1}{n}\sum_{i=0}^{n}\left ( x_i-\hat{x_i} \right )^{2}}
\end{equation}
where, $x$ and $\hat{x}$ represents the vector representation of $SWIR_{24m}$ resampled at 5m GSD and synthesized $SWIR_{5m}$ respectively. In this formulation, we do not apply any form of normalization and report the RMSE as such. Among other evaluation metrics, we use SSIM~\cite{hore2010image}, PSNR~\cite{hore2010image}, SRE~\cite{lanaras2018super} and SAM~\cite{yuhas1992discrimination} indexes to measure reliability of both spatial and spectral profile of the synthesized band.

\subsection{Ablation Study}
\label{ablation}
In order to assimilate the performance of DeepSWIR with different number of convolution layers, we train ablative DeepSWIR with varied number of residual blocks. For faster experimentation, we use Ahmedabad and Kerala regions for training these ablative DeepSWIRs, and Goa for testing its generalization. This training subset is selected so as to cover various objects of interest present in Goa tile. As shown in Table~\ref{ablation_tbl}, both MAE and RMSE in cross-validation follow a decreasing trend as the number of residual blocks (ResBlk), or total convolution operations increases. However, the relative reduction of MAE and RMSE keeps decreasing with depth of the network. Thus, keeping in mind the computational resource, we opt 24 ResBlks with 128 channels each in the customized DeepSWIR. Thereafter, we extend the capacity of the model to incorporate the whole representative dataset by increasing the number of channels/feature size from 128 to 256. The validation MAE and RMSE values of the final model, which we have used in rest of our experimentation, are as low as 4.46 and 7.96, respectively. Note that the ablative trackers have been evaluated on LISS-III dataset by comparing the synthesized $SWIR_{24m}$ with the real $SWIR_{24m}$.
\begin{table}[!th]
\centering
\caption{Comparison of ablative DeepSWIR network architectures.}
\label{ablation_tbl}
\begin{tabular}{|c|c|c|c|c|c|c|}
\hline
\begin{tabular}[c]{@{}l@{}}No. Layer\\ (No. ResBlk)\end{tabular} & \begin{tabular}[c]{@{}l@{}}Feature\\ Size\end{tabular} & Parameters & \multicolumn{2}{|c|}{\begin{tabular}[c]{@{}l@{}}MAE\end{tabular}} & \multicolumn{2}{|c|}{\begin{tabular}[c]{@{}l@{}}RMSE\end{tabular}} \\ \hline
 & & & train & val &  train &  val \\ \hline

15 (6)                                                        & 128                                                    & 1.77M  & 8.64&9.78                                               & 11.56&12.73                                              \\ \hline
27 (12)                                                       & 128                                                    & 3.54M  & 2.64&3.73                                               & 5.70&7.87                                                \\ \hline
35 (16)                                                       & 128                                                    & 4.72M  & 2.09&3.62                                               & 4.54&7.59                                                \\ \hline
51 (24)                                                       & 128                                                    & 7.08M  & 2.01&3.60                                               & 4.34&7.47                                                \\ \hline \hline
51 (24)                                                       & 256                                                    & 28.3M  & 4.13&4.46                                               & 7.27&7.96                                               
\\ \hline
\end{tabular}%

\end{table}

\subsection{Evaluation on LISS-III}
\label{eval3}
\begin{figure}[!th]
	\centering
	\includegraphics[width=0.9\columnwidth]{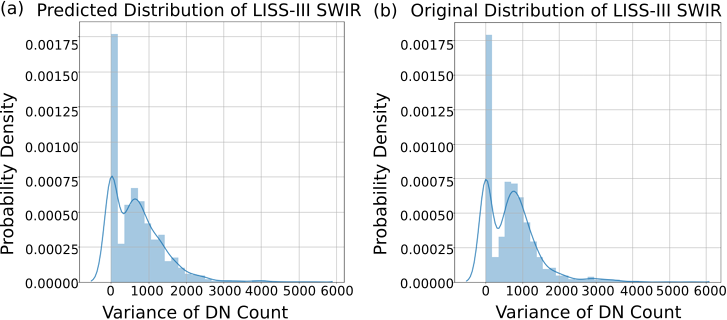}
	\caption{Performance of the model on homogeneous and heterogeneous patches. Variance distribution of (a) LISS-III DeepSWIR and (b) LISS-III SWIR. The histogram of synthesized band follows the histogram of original to a large extent.}
	\label{var_swir}
\end{figure}

Since we train the network on LR bands of LISS-III, we primarily draw a comparison between synthesized  $SWIR_{24m}$ and existing $SWIR_{24m}$. In this regard, the MAE and RMSE errors have been reported in Section~\ref{ablation} and are considerably low in both metrics on approximately $ 200K$ patches covering the aforementioned representative dataset. Additionally, we analyze the network's potential to comprehend the homogeneous and heterogeneous structures in our region of interest. We consider all possible overlapping patches ($[32 \times 32]$) with 50\% overlap along horizontal and vertical direction, and measure their respective variances. As shown in Fig.~\ref{var_swir}, we observed that the synthesized $SWIR_{24m}$ closely resembles the original $SWIR_{24m}$ in terms of homogeneity. This ensures the model's capability to discriminate between homogeneous and heterogeneous textures/features in a tile.

\begin{figure}[!th]
	\centering
	\includegraphics[width=0.9\columnwidth]{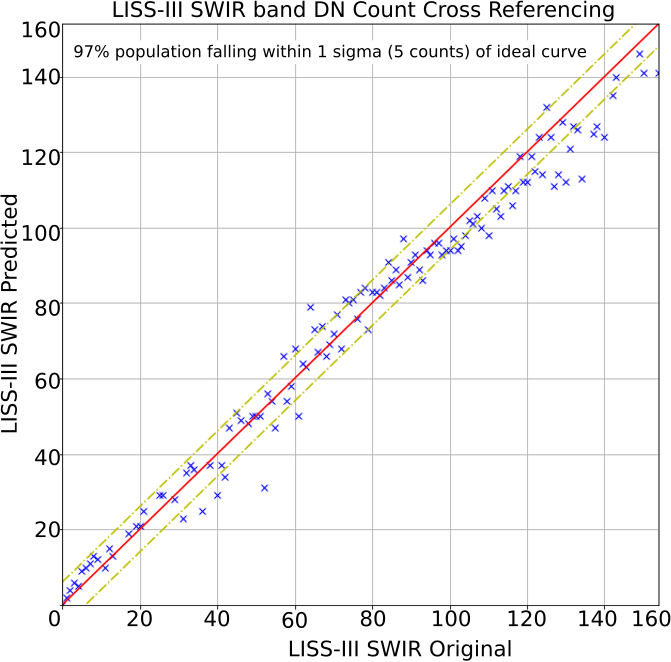}
	\caption{Cross Referencing of predicted and original SWIR band of LISS-III. The predicted DN count closely resembles the original DN count.}
	\label{dn_comp}
\end{figure}

The similarity in variance distribution, however, does not fully characterize the reliability of DN counts. For this reason, we also compare the individual DN counts of synthesized and original $SWIR_{24m}$. As shown in Fig.~\ref{dn_comp}, the scatter plot of DN counts follows the ideal curve (Red) with 97\% population falling within 5 counts of tolerance.  Thus, the combined measure of variance distribution and pixel wise DN counts cross referencing ensures that the model effectively learns spatio-spectral characteristics of individual objects with sufficient degree of realism necessary.

\subsection{Evaluation on LISS-IV}
\label{eval4}
As we focus on synthesizing a high resolution $SWIR_{5m}$, we evaluate the model's transfer learning ability to generalize across different sensors: LISS-III and LISS-IV. Since the model learns to preserve the spatial resolution of input bands through global residual learning during training phase, it synthesizes the virtual band at same spatial resolution as that of the input. Thus, the DeepSWIR model generates a virtual band at 5m GSD when $\left \{ G,R,NIR \right \}_{5m}$ is fed as input. In addition to global residual learning, the DeepSWIR model comprises of local residual learning module, which generates the radiometry of $SWIR$ as a complex combination of $\left \{ G,R,NIR \right \}_{5m}$ and fuses it onto the synthesized virtual band in the high dimensional feature space. The virtual $SWIR_{5m}$, thus synthesized, has been compared with the interpolated $SWIR_{24m}$ at 5m GSD. As given in Table~\ref{l3l4}, the values of state-of-the-art metrics such as RMSE, SSIM, PSNR, SRE, and SAM  between these compared bands are reasonable because of additional structures/details present in synthesized Band5 ($SWIR_{5m}$) which are missing in cubic interpolated $SWIR_{24m}$ at 5m GSD. Nevertheless, the statistics of these state-of-the-art metrics in Band5 are comparable to other bands which verifies the consistency in spatial and spectral characteristics of the synthesized $SWIR_{5m}$.

\begin{table}[!th]
\centering
\caption{Statistics of various state-of-the-art metrics used to evaluate the performance of synthesized SWIR band based on the interpolated LISS-III ${SWIR}_{24m}$ at 5m GSD.}
\label{l3l4}
\begin{tabular}{|l|l|l|l|l|}
\hline 
          & Band2  & Band3  & Band4  & Band5  \\ \hline
RMSE      & 5.057  & 5.565  & 16.488 & 23.367 \\ \hline
SSIM      & 0.989  & 0.979  & 0.898  & 0.879  \\ \hline
PSNR (dB) & 46.305 & 45.804 & 36.725 & 34.723 \\ \hline
SRE (dB)  & 77.279 & 73.464 & 70.919 & 62.263 \\ \hline
SAM (deg) & 2.548  & 5.222  & 5.467  & 10.339 \\ \hline
\end{tabular}%
\end{table}

\begin{figure}[!th] 
	\centering
	\includegraphics[width=0.9\columnwidth]{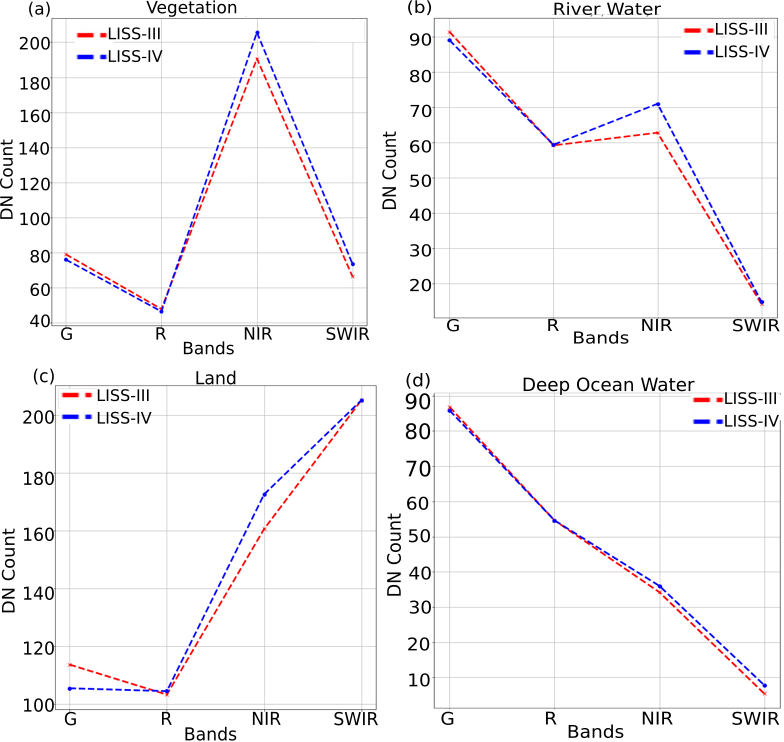}
	\caption{Spectral Response of various terrain features from GOA region (DOP=17.11.17,  path=96, row=62). The synthesized $SWIR_{5m}$ is consistent with other bands of LISS-IV and closely follows the spectral response of LISS-III. }
	\label{sr}
\end{figure}

Further, we analyze the consistency in spectral characteristics of various terrain features in our region of interest as shown in Fig~\ref{sr}. For robust measurement, we consider mean DN count of 10 pixels randomly selected from the compared categories that includes Vegetation (\ref{sr}(a)), River Water (\ref{sr}(b)), Land (\ref{sr}(c)), and Deep Ocean Water (\ref{sr}(d)). As per our experimentation, the spectral characteristics of synthesized band is consistent with existing high resolution bands (LISS-IV) and also, follows a similar pattern in frequency domain as that of the actual low resolution sensor DN counts (LISS-III). 

\subsection{Cross Validation with Sentinel-2B} 
\label{sen}
\begin{figure}
    \centering
    \includegraphics[scale=0.6]{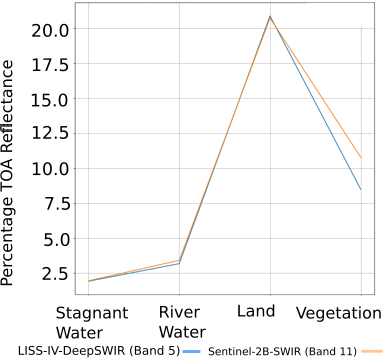}
    \caption{Comparison of percentage TOA reflectance of LISS-IV-DeepSWIR and Sentinel-2B-SWIR. The virtual band closely follows the output of real sensor over homogeneous region.}
    \label{comp}
\end{figure}

Since the model has been trained on LISS-III data only, it may falsely emphasize on learning blind mapping in order to synthesize target band without learning the actual physical characteristics, such as reflectance of various objects. To assimilate such behaviour, we have converted the DN values of $SWIR_{5m}$ to Top Of the Atmosphere (TOA) reflectance and compared with the data product of Sentinel-2B (level 1C) in the nearest wavelength range (Sentinel-2B: (1.64-1.68)$\mu$m, LISS-IV: (1.55-1.70)$\mu$m) and closest date of pass (Sentinel-2B: 01.11.17, LISS-IV: 17.11.17). In order to generate the TOA reflectance, we have used the saturation radiance of LISS-III $SWIR_{24m}$ and DOP of LISS-IV concurrent bands. For robustness, we have considered mean reflectance of 10 randomly selected pixels from the homogeneous regions of chosen categories. Since the native resolution of Sentinel is 20m, we have resampled it to the resolution of LISS-IV (5m) using cubic interpolation. As shown in Fig.~\ref{comp}, the TOA reflectance of the virtual sensor: DeepSWIR closely follows that of the real sensor: Sentinel-2B over homogeneous regions: Stagnant Water, River Water, and Land. The relatively smaller TOA reflectance of Vegetation regions of LISS-IV is due to additional structures/details present in our DeepSWIR which has native resolution of 5m unlike Sentinel-2B.

\subsection{Application of Synthesized Band in Wetland Inventory}
\label{wetland}

\begin{figure}
    \centering
    \includegraphics[width=0.9\columnwidth]{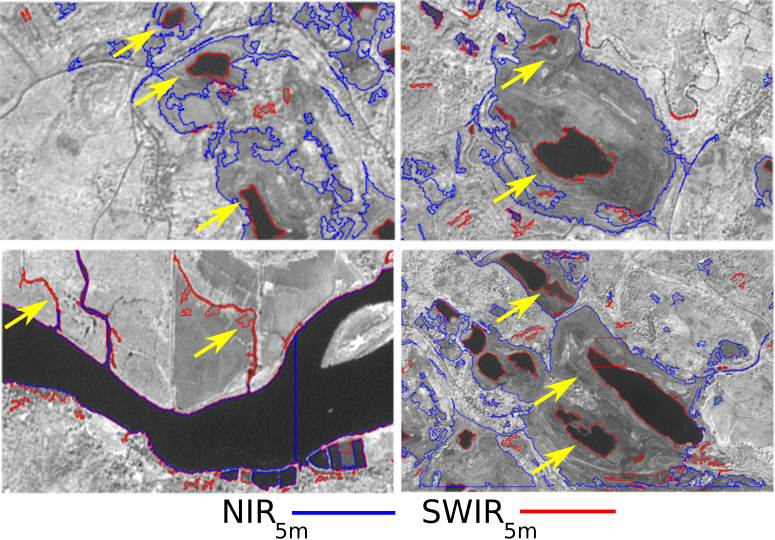}
    \caption{Comparison of $NIR_{5m}$ and synthesized $SWIR_{5m}$ in wetland mapping. Unlike $NIR_{5m}$, the synthesized high resolution $SWIR_{5m}$ assists in delineating narrow wetland resources and suppress false positives to a great extent.}
    \label{wet}
\end{figure}

\begin{figure*}[!th]
    \centering
    \includegraphics[scale=0.7]{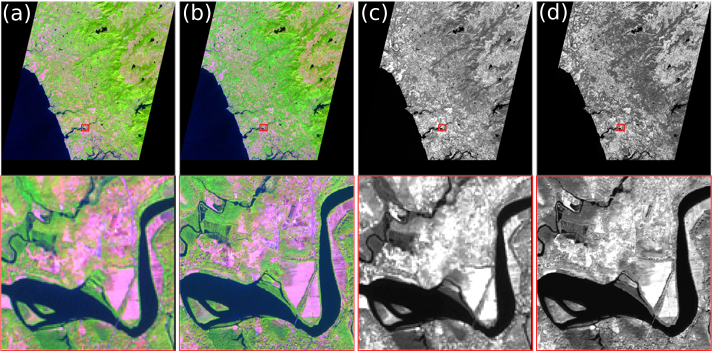}
    \caption{(a) FCC of LISS-III $\left\lbrace SWIR,NIR,R \right\rbrace_{24m}$ resampled at 5m. (b) FCC of LISS-IV $\left\lbrace SWIR(Synthesized),NIR,R \right\rbrace_{5m}$. (c) LISS-III $SWIR_{24m}$ resampled at 5m. (c) LISS-IV $SWIR_{5m}$ Synthesized. The bottom row represents the selected region highlighted by red bounding box. The FCCs ensure the consistency in both radiometric resemblance (a-b) and geometric  registration of synthesized band with respect to existing high resolution bands (b). The radiometric resemblance between SWIR interpolated and synthesized band (c-d) verifies enhanced spatial resolution with desired spectral characteristics. }
    \label{l3l4_full1}
\end{figure*}

In order to verify the usefulness of synthesized band in real world applications, we have used the generated $SWIR_{5m}$ to delineate wetland resources in selective regions of GOA tile (path=96, row=62). Since our primary scope of this study is not to develop wetland delineation algorithm, but synthesis of high resolution $SWIR_{5m}$ band, we have used a simple and yet very effective algorithm proposed by Sushma et. al.~\cite{panigrahy2012wetlands} for mapping wetland resources. As shown in Fig.~\ref{wet}, the use of high resolution $NIR_{5m}$ in the absence of $SWIR_{5m}$ fails to generate accurate maps of wetland resources and fails to delineate narrow wetlands as higlighted by the arrow heads. On the contrary, the use of synthesized $SWIR_{5m}$ could capture the unique features associated with various classes of wetland resources and hence, could suppress false positives and delineate even narrow wetlands to a great extent. In Fig.~\ref{l3l4_full1} we show qualitative comparison between original and synthesized band where minute structures/details present in synthesized band are clearly distinguishable from the interpolated band at 5m GSD.

%

\section{Concluding Remarks}
\label{Conclusion}
In this study, we demonstrated an efficient way to utilize the power of deep learning techniques in synthesizing a virtual band using existing bands of multi-sensor satellite imagery. Further, we critically analyzed the qualitative and quantitative results produced by the proposed DeepSWIR with extensive experimentation and appropriate evaluation standards. Additionally, we proposed a Gaussian feathering based image stitching mechanism for seamless band generation. As per our experiments, we report that the synthesized high resolution $SWIR_{5m}$ possesses the spectral characteristics of LISS-III $SWIR_{24m}$ and spatial characteristics of LISS-IV sensor to a great extent. We also observed a significant resemblance in the TOA reflectance of the virtual band, $SWIR_{5m}$ with that of the real sensor, Sentinel-2B over homogeneous features. At the end, we showed the real world application of the synthesized high resolution band by using it to delineate wetland resources over our region of interest. \\ 

 However, though the computationally extensive deep convolutional neural networks are highly efficient in synthesizing the high resolution virtual band, $SWIR_{5m}$ that closely mimics the spatio-spectral characteristics of real bands, this method should not be considered as an alternative to sending remote sensing satellites. First, these models are data driven approaches and hence, strongly depend upon representative training datasets which are acquired by remote sensing satellites. Second, the model extracts spatial information from the existing high resolution concurrent bands: $\left \{ G,R,NIR \right \}_{5m}$, and spectral information from existing low resolution bands: $\left \{ G,R,NIR,SWIR\right \}_{24m}$. This model can generate a high resolution band at one particular time provided it has access to other high resolution accurate bands, $\left \{G,R,NIR\right \}_{5m}$ at that instant of time. Therefore, unlike real SWIR band sensor, the DeepSWIR model can not produce $SWIR_{5m}$ in highly unlikely scenarios where all its source band sensors fail. Nevertheless, our study shows that a deep neural network can be successfully employed to synthesize virtual bands in most cases and more importantly, it shows appealing results in both qualitative and quantitative assessments. Therefore, these biologically inspired techniques can be used to broaden our vision in the EMS with certain degree of accuracy when technology reaches its limit.


%

%
%

\section*{Acknowledgment}
We are immensely grateful to our colleagues Ashwin Gujrati and Vibhuti Bhushan Jha who provided expertise that greatly assisted the research, although any errors are our own and should not tarnish the reputation of these esteemed professionals. We would also like to express our gratification to all the members of Optical Data Processing Division (ODPD), Signal and Image Processing Group (SIPG) for their continuous support throughout this research.
\ifCLASSOPTIONcaptionsoff
  \newpage
\fi



%

\bibliographystyle{ieeetr}
\bibliography{egbib}
\end{document}